# A Machine Learning Approach for the Identification of Bengali Noun-Noun Compound Multiword Expressions

**Vivekananda Gayen**
Central Calcutta Polytechnic
Kolkata – 700014, India
vivek3gayen@gmail.com

**Kamal Sarkar**
Computer Science & Engineering Dept.
Jadavpur University
Kolkata-700032, India
jukamal2001@yahoo.com

**Abstract**

This paper presents a machine learning approach for identification of Bengali multiword expressions (MWE) which are bigram nominal compounds. Our proposed approach has two steps: (1) candidate extraction using chunk information and various heuristic rules and (2) training the machine learning algorithm called *Random Forest* to classify the candidates into two groups: bigram nominal compound MWE or not bigram nominal compound MWE. A variety of association measures, syntactic and linguistic clues and a set of WordNet-based similarity features have been used for our MWE identification task. The approach presented in this paper can be used to identify bigram nominal compound MWE in Bengali running text.

## 1 Introduction

Baldwin and Kim (2010) defined multiword expressions (MWEs) as lexical items that: (a) can be decomposed into multiple lexemes; and (b) display lexical, syntactic, semantic, pragmatic and/or statistical idiomaticity.

Most real world NLP applications tend to ignore MWE, or handle them simply by listing, but successful applications will need to identify and treat them appropriately.

Automatic identification of multiword expression (MWE) from a text document can be useful for many NLP (natural language processing) applications such as information retrieval, machine translation, word sense disambiguation.

In terms of the semantics, compositionality is an important property of MWEs. Compositionality is the degree to which the features of the parts of a MWE combine to predict the features of the whole. According to the compositionality property, the MWEs can take a variety of forms: complete compositionality (also known as institutionalized phrases (e.g. *rAjya sarkAr* (state government)), partial compositionality (e.g. *Am Admi* (common people)), idiosyncratic compositionality (e.g. spill the beans) and finally complete non-compositionality (e.g. *ubhoy sangkat* (on the horns of a dilemma)).

Since compound nouns are productive and new compound nouns are created from day to day, it is impossible to exhaustively store all compound nouns in a dictionary. It is also common practice in Bengali literature to use MWEs which are compound nouns. Bengali new terms directly coined from English terms are also commonly used as MWEs in Bengali (e.g. *nano sim*). The primary types of noun-noun MWEs in Bengali are: Named-Entities (NE) (name of a person, an organization, a location etc.), Idiomatic Compound Nouns ( e.g., *kal kArkhAnA* (mills and workshops)), Idioms(e.g., *tAser ghar* (any construction that may tumble down easily at any time)), Numbers(e.g., *soyA teen ghantA* (three hours and fifteen minutes)), Relational Noun Compounds(e.g., *mejo meye* (second daughter)), Conventionalized Phrases (e.g., *chAkkA jyAm* (standstill)), Simile terms (e.g., *hAter pAnch* (last resort)), Reduplicated terms (e.g., *bAri bAri*

*A Machine Learning Approach for the Identification of Bengali Noun-Noun Compound Multiword Expressions* 291(door to door)), Administrative terms(e.g., *sarAstrA montrak* (home ministry)), phrases with one of the components coined from English literature(e.g., *mAdrAshA board*), phrases with both of the components coined from English literature(e.g., *roaming chArge*).

Multiword expression extraction approaches can be broadly classified as: association measure based methods, deep linguistic based methods, machine learning based methods and hybrid methods.

The earliest works on MWE extraction used statistical measures for multiword expression extraction. The system called Xtract (Smadja, 1993) uses positional distribution and part-of-speech information of surrounding words of a word in a sentence to identify interesting word pairs. Classical statistical hypothesis tests like Chi-square test, t-test, z-test, log-likelihood ratio (Manning and Schütze, 2000), mutual information and point-wise mutual information (Bouma, 2009) have also been employed to extract collocations.

Kunchukuttan and Damani (2008) used various statistical measures such as point-wise mutual information, log-likelihood, frequency of occurrence for extraction of Hindi compound noun multiword expression. Agarwal et al. (2004) has used co-occurrence and significance function to extract MWE automatically in Bengali, focusing mainly on noun-verb MWE. Chakraborty (2010) has used a linear combination of association measures namely co-occurrence, Phi, significance function to obtain a linear ranking function for Bengali noun-noun collocation candidates and MWEness is measured by the rank score assigned by the ranking function.

Piao et al. (2005) focuses on symbolic approach to multiword extraction that uses large-scale semantically classified multiword expression template database and semantic field information assigned to MWEs by the USAS semantic tagger. Sinha (2011) has used a stepwise methodology that exploits linguistic knowledge such as replicating words, pair of words, samaas, sandhi and vaalaa morpheme constructs for mining Hindi MWEs. A Rule-Based approach for identifying only reduplication from Bengali corpus has been presented in (Chakraborty and Bandyopadhyay, 2010). A semantic clustering based approach for identifying bigram noun-noun MWEs from a medium-size Bengali corpus has been presented in (Chakraborty et al., 2011).

Pecina (2008) used linear logistic regression, linear discriminant analysis (LDA) and Neural Networks separately on feature vector consisting of 55 association measures for extracting MWEs. Venkatapathy et al. (2005) has presented an approach to measure relative compositionality of Hindi noun-verb MWEs using Maximum entropy model.

Hybrid method combines statistical, linguistic and/or machine learning methods. Maynard and Ananiadou (2000) combined both linguistics and statistical information in their system, TRUCK, for extracting multiword terms. Dias (2003) has developed a hybrid system for MWE extraction, which integrates word statistics and linguistic information. Ramisch et al. (2010) presents a hybrid approach to multiword expression extraction that combines the strengths of different sources of information using a machine learning algorithm.

The main focus of our work is to develop a machine learning approach that uses a set of statistical, syntactic and linguistic features for identifying Bengali multiword expressions (MWE) which are bigram nominal compounds. To date, not much comprehensive work has been done on Bengali multiword expression identification. Very recently, Gayen and Sarkar (2013) uses random forest that uses some association measures and some syntactic features for noun-noun MWE identification. We have compared the performance of the system presented in (Gayen and Sarkar, 2013) to our proposed system presented in this paper.

The proposed noun-noun MWE identification method has been detailed in section 2. The evaluation and results are presented in section 3.

## 1 Proposed Noun-Noun MWE Identification Method

Our proposed noun-noun MWE identification method has several steps: preprocessing, candidate extraction and noun-noun MWE identification by classifying the candidates into two categories: positive (noun-noun MWE) and negative (not noun-noun MWE).

## 1.1 Preprocessing

At this step, unformatted documents are segmented into a collection of sentences automatically according to Dari (in English, full stop), Question mark (?) and Exclamation sign (!). Then the sentences are submitted to the chunker (**http//ltrc.iiit.ac.in/analyzer/bengali**) one by one for processing. The chunked output is then processed to delete the information which is not required for MWE candidate identification task.

## 1.2 Candidate Noun-Noun MWE Extraction

The chunked sentences are processed to identify the noun-noun multi-word expression candidates. The multiword expression candidates are primarily extracted using the following rule found in (Bharati et al., 2006):

Bigram consecutive noun-noun token sequence (except binary number expression) within same NP (Noun Phrase) chunk is extracted from the chunked sentences if the tag of the token is NN or NNP or XC (NN: Noun, NNP: Proper Noun, XC: compounds).

It is observed that some potential candidates are missed due to chunkers error but more number of potential noun-noun MWE candidates are identified from the unchunked corpus using the following heuristics:

*Bigrams which are hyphenated or reduplicated or occur within single quote or within first brackets or whose words are out of vocabulary (OOV) are also considered as the potential candidates for noun-noun MWE. Binary number expression candidates are not considered here.*

## 1.3 Features

**Statistical features**: Absolute frequencies are in no way able to capture the associations of words forming a MWE. The alternative to relying on absolute frequencies is to use a statistical association measure like the Mutual Information (MI) score. Association measure scores reflect the collocation strength of pairs of words forming the bigram MWE. As the individual word frequencies become higher, it becomes more likely that the word combination would occur just by random chance, and therefore the combination has less importance. We consider a number of association measures as the statistical features.

We use the association measures namely phi, point-wise mutual information (pmi), salience, log likelihood, poisson stirling, chi, t-score, co-occurrence and significance to calculate the scores of each noun-noun candidate MWE. The detail of these statistical features namely phi, point-wise mutual information (pmi), salience, log likelihood, Poisson stirling, chi, t-score can be found in (Banerjee and Pedersen, 2003; Gayen and Sarkar, 2013) and details of co-occurrence and significance can be found in (Agarwal et al., 2004; Gayen and Sarkar, 2013). These association measures use various types of frequency statistics associated with the bigram. Since Bengali is highly inflectional language, the noun-noun candidate MWEs are stemmed while their frequencies are computed.

In our work, we use a lightweight stemmer for Bengali that strips the suffixes using a predefined suffix list, on a "longest match" basis, using the algorithm similar to that for Hindi (Ramanathan and Rao, 2003).

**WordNet similarity features**: We use an open source WordNet::Similarity package (Pedersen et al., 2004) to compute semantic similarity or relatedness between a pair of components of a candidate noun-noun compound MWE. We translate the components of Bengali noun-noun MWE candidate using a bilingual dictionary and then submit the translated pair to WordNet::Similarity program to get similarity score. During word translation first translation for the word in the dictionary is considered. Similarity measures that we use are: Lin (the Lin measures), wup (the Wu and Palmer measure), path (simple node counts (inverted)), vector (gloss vector measure) and vector-pairs (pairwise gloss vector measure).

**Syntactic and linguistic features**: Other than the statistical and WordNet based features discussed in the above subsections, we also use some syntactic and linguistic features which are listed in the table 1.

| Feature name | feature description | Feature type |
|---|---|---|
| Avg-WordLength | average length of the components of a candidate MWE | Continuous |
| Whether-Hyphenated | Whether a candidate MWE is hyphenated | Binary |
| Whether-Within-Quote | Whether a candidate MWE is within single quote | Binary |
| Whether-Within-Bracket | Whether a candidate MWE is within first brackets | Binary |
| OOV | Whether candidate MWE is out of vocabulary | Binary |
| Reduplication | Whether candidate is reduplicated | Binary |
| First-Word-Inflection | Whether the first word is inflected | Binary |
| Second-Word-Inflection | Whether second word is inflected | Binary |
| Tag-Of-FirstWord | Lexical category of the first word of a candidate | Nominal (values: XC, NN, NNP) |
| Tag-Of-SecondWord | Lexical category of the second word of a candidate | Nominal (values: XC, NN, NNP) |

Table 1: Syntactic and linguistic features

## 1.4 Noun-noun MWE Identification using Random Forest

Random forest (Breiman, 2001) is an ensemble classifier that combines the predictions of many decision trees using majority voting to output the class for an input vector. Each decision tree participated in ensembling chooses a subset of features randomly to find the best split at each node of the decision tree.

The method combines the idea of "bagging" (Breiman, 1996) and the random selection of features. We use this algorithm for our multiword identification task for several reasons: (1) For many data sets, it produces a highly accurate classifier (Caruana et al, 2008 ), (2) It runs efficiently on large databases and performs well consistently across all dimensions and (3) It generates an internal unbiased estimate of the generalization error as the forest building progresses. The outline of the algorithm is given in the figure 1.

Training Random Forests for noun-noun MWE identification requires candidate noun-noun MWEs to be represented as the feature vectors. For this purpose, we write a computer program for automatically extracting values for the features characterizing the noun-noun MWE candidates in the documents. For each noun-noun candidate MWE in a document in our corpus, a feature vector is constructed using the values of the features of the candidate. If the noun-noun candidate MWE is found in the list of manually identified noun-noun MWEs, we label the corresponding feature vector as "Positive" and if it is not found we label it as a "negative". Thus the feature vector for each candidate looks like { < $a_1$ $a_2$ $a_3$ ….. $a_n$>, <label> } which becomes a training instance (example) for the random forest, where $a_1$, $a_2$ . . .$a_n$, indicate feature values for a candidate. Our training data consists of a set of instances of the above form.

For our experiment, we use Weka (www.cs.waikato.ac.nz/ml/weka) machine learning tools. The random forest is included under the panel Classifier/ trees of WEKA workbench.. For our work, the random forest classifier of the WEKA suite has been run with the default values of its parameters. One of the important parameters is the number of trees used for building the forest. We set this parameter to its default value of 10.

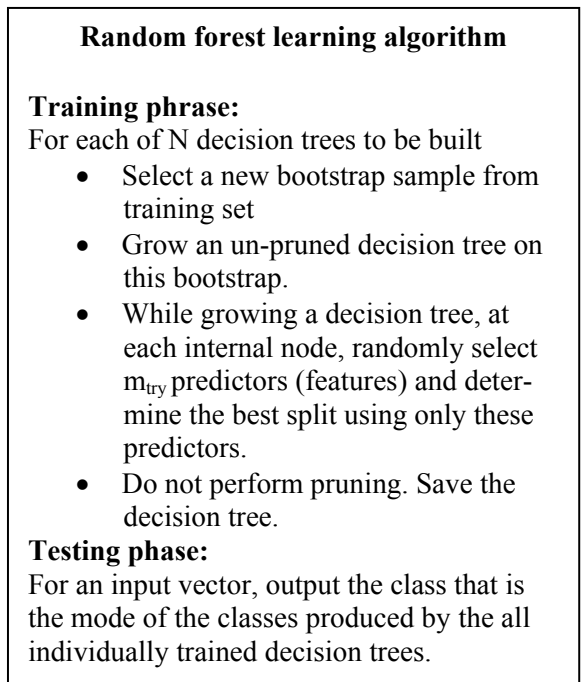

Figure 1: Random forest learning algorithm

## 2 Evaluation, Experiments and Results

For evaluating the performance of our system the traditional precision, recall and F-measure are computed by comparing machine assigned labels to the human assigned labels for the candidate noun-noun MWEs.

For system comparisons, 10-fold cross validation is done to estimate overall accuracy of each of the systems presented in this paper. The dataset is randomly reordered and then split into *n* parts of equal size. For each of 10 iterations, one part is used for testing and the other *n*-1 parts are used for training the system. The test results are collected and averaged over all folds. This gives the cross-validation estimate of the accuracy of the system. For evaluating the performance of our system the traditional F-measure is computed in terms of precision, recall by comparing machine assigned labels to the human assigned labels for the candidate noun-noun MWEs.

### 2.1 Experimental Dataset

For our experiments, we have created a corpus by collecting the news articles from the online version of well known Bengali newspaper ANANDABAZAR PATRIKA during the period spanning from 20.09.2012 to 19.10.2012. It consists of total 274 documents and all those documents contain 18769 lines of Unicode texts, 233430 tokens.

We have manually identified the noun-noun compound MWEs in the collection and prepared the training data by assigning positive labels to the noun-noun MWEs and negative labels to the expressions which are not noun-noun MWEs. While the noun–noun multiword expressions are manually identified by us from the corpus the following criteria are taken into consideration:

(1) Whether a noun-noun sequence is institutionalized by usages
(2) Whether a noun-noun sequence is partially or completely non-compositional

If any of the above mentioned criteria is satisfied by a noun-noun sequence, we have considered it as a noun-noun MWE.

Finally we have manually identified from our corpus 4664 noun-noun MWEs of two types: bigram noun-noun compounds and named entities.

Total 8546 candidate noun-noun MWEs are automatically extracted by employing chunker and heuristic rules as described in subsection 2.2.

### 2.2 Experiments

We conducted several experiments to judge the effectiveness of the proposed system.

**Experiment 1**: For this experiment, we have developed our proposed system using a combined set of statistical, syntactic, linguistic and WordNet based similarity features discussed earlier in this paper.

**Experiment 2**: This is to design a baseline system (*baseline 1*) to which the proposed system is compared. Since Chakraborty (2010) has used a linear combination of several association measures to obtain a linear ranking function for Bengali noun-noun collocation candidates, we have designed a baseline system which considers only the association measures namely phi, point-wise mutual information (PMI), salience, log likelihood,

Poisson stirling, chi, t-score, co-occurrence and significance. Unlike the work presented in Chakraborty (2010) that uses manual tuning of weights to obtain a linear ranking function, we use the machine learning algorithm called Random Forest for feature combination.

**Experiment 3**: In this experiment, we design another baseline system (*baseline 2*) that uses only WordNet based similarity features discussed earlier in this paper. We have designed this baseline with the WordNet based similarity features, because the work presented in (Chakraborty et al., 2011) has used WordNet based similarity features to detect Bengali noun-noun MWEs. Here we consider the various WordNet based similarity measures namely Lin, wup, path, vector and vector-pairs as the features and use random forest for feature combination.

**Experiment 4**: In this experiment, we compare the proposed method to the noun-noun MWE identification method presented in (Gayen and Sarkar, 2013). Though the approach presented in (Gayen and Sarkar, 2013) also uses random forest for noun-noun MWE identification, our proposed approach incorporates some additional features such as reduplication and WordNet based similarity features which are not considered in (Gayen and Sarkar, 2013). We consider the system presented in (Gayen and Sarkar, 2013) as the baseline system 3.

## 2.3 Results

The comparisons of the performance of our proposed system with the different baseline systems in terms of the weighted average F-measures are shown in table 2. F-measure shown in the table is the weighted (by class size) average of F-measures achieved by a system for the different output classes. For our case, there are two output classes: noun-noun MWE (positive) and not noun-noun MWE (negative). F-measure for an output class is computed by combining traditional precision and recall measures using the formula: $(2*P*R) / (P + R)$, where P and R are respectively the precision and recall achieved by a system for a particular class.

Table 2 shows that our proposed system, which uses a combined set of statistical, syntactic, linguistic and WordNet based similarity features, performs better than other three baseline systems to which the proposed system is compared.

| Systems | F-measure (weighted average) |
|---|---|
| Our proposed system | **0.869** |
| Baseline system 3 (The system presented in (Gayen and Sarkar, 2013)) | 0.852 |
| Baseline system1(the system with association measure based features) | 0.804 |
| Baseline system 2 (system with only WordNet based similarity features) | 0.673 |

Table 2: Comparisons of the performances of our proposed system with the baseline systems.

## 3 Conclusion

This paper presents a machine learning based approach for identifying noun-noun compound MWEs from the Bengali corpus. We have used a number of association measures, syntactic and linguistic information as features which are combined by a random forest learning algorithm for recognizing noun-noun compound MWEs. The approach presented in this paper can be used to identify noun-noun MWEs in Bengali running text. With the suitable modifications in the feature set, the approach presented in this paper can be applied to identification of other types of MWEs from the Bengali corpus.

As a future work, we have planned (1) to improve the system performance by improving candidate MWE extraction step of the proposed system and/or introducing more number of new features, and (2) to apply the proposed approach to identification of other types of Bengali MWEs.